\documentclass[lettersize,journal]{IEEEtran}
\usepackage{amsmath,amsfonts}
\usepackage{algorithm}
\usepackage{algpseudocode}
\usepackage{array}
\usepackage[caption=false,font=normalsize,labelfont=sf,textfont=sf]{subfig}
\usepackage{textcomp}
\usepackage{stfloats}
\usepackage{url}
\usepackage{verbatim}
\usepackage{graphicx}
\usepackage{cite}
\usepackage{color}
\usepackage{xcolor}
\usepackage{bm}
\usepackage{amsmath}
\usepackage{amssymb}
\usepackage{multirow}
\usepackage{hyperref}
\usepackage{booktabs}
\usepackage{colortbl}
\usepackage{orcidlink} 
\usepackage{bbm}

\definecolor{LightRed}{rgb}{1,0.92,0.92}
\definecolor{LightOrange}{rgb}{1,0.95,0.88}
\definecolor{LightYellow}{rgb}{1.0,1.0,0.84}
\definecolor{LightGreen}{rgb}{0.9,1.0,0.88}
\definecolor{LightCyan}{rgb}{0.9,1,1}
\definecolor{LightBlue}{rgb}{0.9,0.94,1}

\begin{document}
\title{HCVP: Leveraging Hierarchical Contrastive Visual Prompt for Domain Generalization}
\author{Guanglin~Zhou\orcidlink{0000-0002-1305-2622}\thanks{This work was conducted during Guanglin Zhou's internship at Mohamed bin Zayed University of Artificial Intelligence (MBZUAI).},
    Zhongyi~Han\orcidlink{0000-0003-2851-193X},
    Shiming~Chen\orcidlink{0000-0001-9633-3392},
    Biwei~Huang,
    Liming~Zhu,
    Tongliang~Liu\orcidlink{0000-0002-9640-6472},~\IEEEmembership{Senior~Member,~IEEE},
    Lina~Yao\orcidlink{0000-0002-4149-839X},~\IEEEmembership{Senior~Member,~IEEE}, 
    Kun~Zhang
\IEEEcompsocitemizethanks{
\IEEEcompsocthanksitem Guanglin Zhou is with the University of New South Wales.
Email: jameszhou.ustc@gmail.com (corresponding author).
\IEEEcompsocthanksitem Zhongyi Han and Shiming Chen are with Carnegie Mellon University and Mohamed bin Zayed University of Artificial Intelligence. Email: hanzhongyicn@gmail.com, gchenshiming@gmail.com.
\IEEEcompsocthanksitem Biwei Huang is with University of California, San Diego. Email: bih007@ucsd.edu.
\IEEEcompsocthanksitem Liming Zhu is with CSIRO's Data61. Email: liming.zhu@data61.csiro.au.
\IEEEcompsocthanksitem Tongliang Liu is with the University of Sydney and Mohamed bin Zayed University of Artificial Intelligence. Email: tongliang.liu@sydney.edu.au.
\IEEEcompsocthanksitem  Lina Yao is with CSIRO's Data61, the University of New South Wales and Macquarie University. Email: lina.yao@unsw.edu.au.
\IEEEcompsocthanksitem Kun Zhang is with Carnegie Mellon University and Mohamed bin Zayed University of Artificial Intelligence. Email: kunz1@cmu.edu.
}
}

\markboth{Journal of \LaTeX\ Class Files,~Vol.~14, No.~8, August~2021}%
{Shell \MakeLowercase{\textit{et al.}}: A Sample Article Using IEEEtran.cls for IEEE Journals}

\maketitle

\begin{abstract}
Domain Generalization (DG) endeavors to create machine learning models that excel in unseen scenarios by learning invariant features. 
In DG, the prevalent practice of constraining models to a fixed structure or uniform parameterization to encapsulate invariant features can inadvertently blend specific aspects.
Such an approach struggles with nuanced differentiation of inter-domain variations and may exhibit bias towards certain domains, hindering the precise learning of domain-invariant features.
Recognizing this, we introduce a novel method designed to supplement the model with domain-level and task-specific characteristics.
This approach aims to guide the model in more effectively separating invariant features from specific characteristics, thereby boosting the generalization.
Building on the emerging trend of visual prompts in the DG paradigm, our work introduces the novel \textbf{H}ierarchical \textbf{C}ontrastive \textbf{V}isual \textbf{P}rompt (HCVP) methodology.
This represents a significant advancement in the field, setting itself apart with a unique generative approach to prompts, alongside an explicit model structure and specialized loss functions.
Differing from traditional visual prompts that are often shared across entire datasets, HCVP utilizes a hierarchical prompt generation network enhanced by prompt contrastive learning. 
These generative prompts are instance-dependent, catering to the unique characteristics inherent to different domains and tasks.
Additionally, we devise a prompt modulation network that serves as a bridge, effectively incorporating the generated visual prompts into the vision transformer backbone.
Experiments conducted on five DG datasets demonstrate the effectiveness of HCVP, outperforming both established DG algorithms and adaptation protocols. The code is available at: \url{https://github.com/jameszhou-gl/TMM-HCVP}.
\end{abstract}


\begin{figure}[!tb]
\centering
\includegraphics[width=1.0\linewidth]{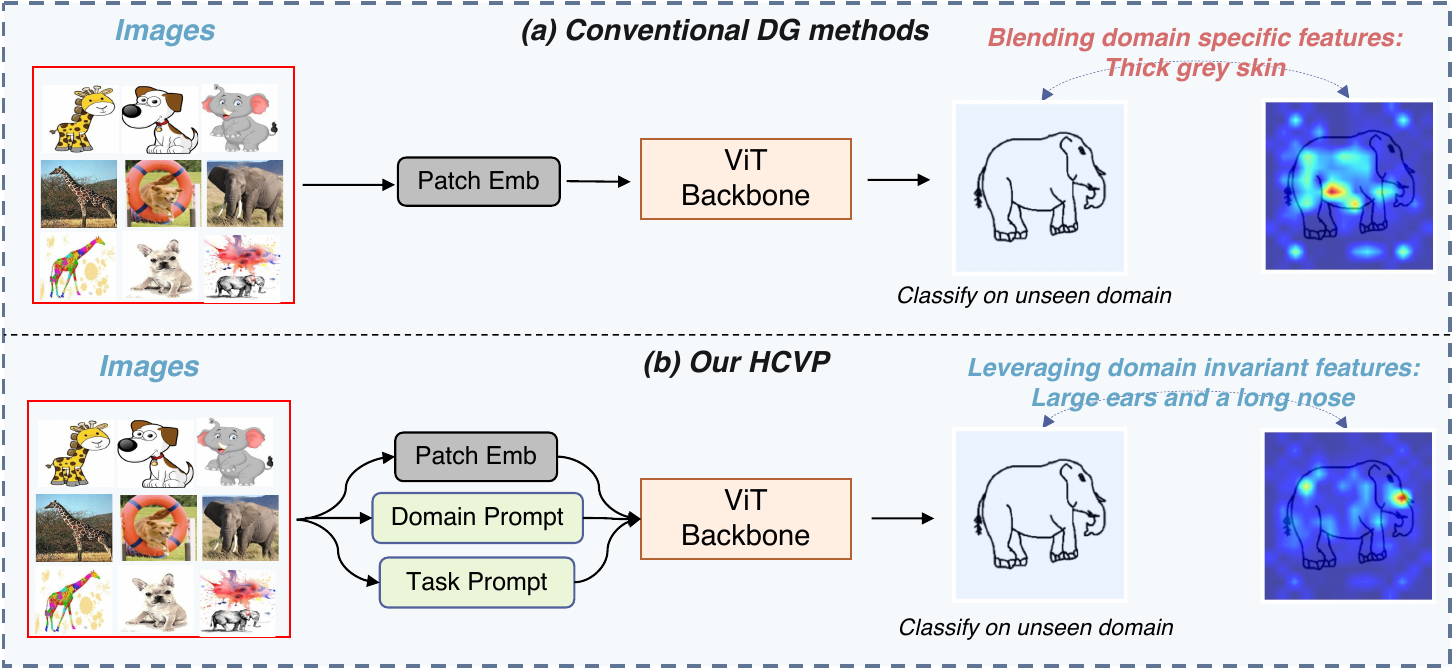}
\caption{Motivation illustration.
(a) Traditional DG methods, employing universal parameters across the entire dataset, often struggle to distinguish between invariant shape attributes (e.g., large ears, long nose) and domain-specific texture attributes (e.g., thick grey skin).
This difficulty leads to a blending of features, thereby diminishing the model's capacity for generalization. 
(b) Our approach introduces hierarchical visual prompts that encapsulate domain-level and task-specific characteristics, enabling the model to better differentiate and understand both invariant and specific attributes, thereby contributing to more effective generalization across different domains.}
\label{266492058979}
\end{figure} 

\section{Introduction}
Despite the remarkable successes of machine learning models, particularly deep neural networks (DNNs), in various areas, they often exhibit unexpected failures when the test data distribution deviates from the training data~\cite{wang2022generalizing,zhou2022domain,zhou2023emerging}.
For instance, adversarial inputs can lead to misclassifications \cite{goodfellow2014explaining}, performance on digit recognition may deteriorate under rotations \cite{piratla2020efficient}, and DNNs trained to recognize pneumonia may fail when applied to scans from new hospitals \cite{geirhos2020shortcut}.
Addressing this, domain generalization (DG) seeks to create robust models that can effectively adapt and perform on unseen, yet related, domains by leveraging invariant features learned across multiple source domains \cite{wang2022generalizing,zhou2022domain}. 

Domain generalization has traditionally relied on techniques such as data manipulation and representation learning \cite{wang2022generalizing,zhou2022domain}, giving rise to classic DG algorithms \cite{ganin2016domain,rojas2018invariant, sun2021recovering, christiansen2021causal, li2018deep, wang2023causal}. 
Despite these efforts, several large-scale empirical studies \cite{gulrajani2020search,wiles2022a} have highlighted limited performance improvements in current DG algorithms. 
Alongside these traditional methods, recent advancements in visual foundation models \cite{bommasani2021opportunities}, such as vision transformer (ViT) backbone~\cite{dosovitskiy2021an}, have been leveraged for DG tasks~\cite{zhang2022delving,chen2023explore}. 
Their ability to capture a broad spectrum of diverse features has opened up new avenues for effective DG.
Several tuning protocols have been proposed to adapt these foundation models to downstream tasks \cite{kumar2022fine,li2022sparse,rame2023model,du2021fewshot,kumar2022fine,lee2022surgical}.

The fundamental objective in DG lies in effectively discerning and leveraging domain-invariant features from specific characteristics.
Nevertheless, the challenge lies in accomplishing this separation within a unified modeling framework without extra, tailored information.  
The inherent complexity of handling diverse and often subtle differences between domains necessitates additional context beyond what a single model can provide.
This lack of specificity leads to a DG approach that unintentionally blends domain-specific aspects, making it challenging to differentiate nuanced inter-domain variations \cite{geirhos2020shortcut}.
For example, the DG dataset depicted in Figure \ref{266492058979}, encompasses the classification tasks like classifying elephants across diverse domains such as the cartoon, photo, and sketch.  
Shape attributes like large ears and texture attributes such as thick grey skin are crucial for prediction. 
However, the significance of the thick grey skin might diminish across domains. 
Without specific adjustments, a conventional model might struggle to discern these features, leading to the blending of domain-specific attributes in predictions on unseen domains. 
This complexity highlights the need for a more nuanced approach to handle the unique challenges.

In this work, we introduce a novel method that supplements raw images with domain-level and task-specific characteristics, aiming to guide the model in better learning invariant features from specific aspects.
Such insight paves the way for our exploration of visual prompts, which have served as contextual anchors, enriching models with a tailored understanding within specific contexts \cite{radford2021learning,zhou2022learning,jia2022visual}. 
Building on the emerging trend of visual prompts in the DG paradigm~\cite{zheng2022prompt,li2022learning}, we introduce the novel \textbf{H}ierarchical \textbf{C}ontrastive \textbf{V}isual \textbf{P}rompt (HCVP) methodology, representing a significant advancement in this field.
Differing from traditional visual prompts that are often shared across the entire dataset, HCVP employs a two-tier hierarchical prompt generation network along with a prompt contrastive learning strategy \cite{chen2020simple,zhou2023contrastive,li2022contrastive}.
Such explicit model structure and specialized loss functions guide the visual prompt learning process to capture both domain-level and task-specific details. 
Moreover, we devise a prompt modulation network that acts as a bridge, effectively incorporating our generated visual prompts into the ViT backbone.
In order to enhance visual features with the discriminative power between inter-classes, we introduce class-conditioned contrastive invariance. 
The experiments, conducted on several DG datasets \cite{gulrajani2020search,ye2022ood}, demonstrate the effectiveness of our approach.   

The main contributions of this paper are summarized as: 
\begin{itemize}
    \item We introduce a novel approach that integrates domain and task information in the DG process, enhancing the distinction of invariant features and those specific to individual domains.
    \item Our HCVP method innovatively utilizes generative visual prompts tailored for DG, explicitly employing a hierarchical prompt generation network and prompt contrastive learning, to craft prompts that encapsulate relevant domain-level and task-specific characteristics.
    \item Through extensive experiments, HCVP is demonstrated to achieve state-of-the-art performance on five real-world datasets in DG, surpassing both established DG algorithms and adaptation protocols.
\end{itemize}

\section{Related Work}
\subsection{Domain Generalization}
Domain generalization traditionally relied on data manipulation and representation learning \cite{wang2022generalizing}. Techniques such as data augmentation \cite{yue2019domain} and generation \cite{qiao2020learning} were utilized to diversify training data. 
Representation learning~\cite{jo2023poem,cha2021swad,cha2022domain,liu2021domain,jin2021style,ma2019deep,luo2023taking,niu2023knowledge,wang2021learning,gokhale2023improving} used strategies ranging from feature alignment \cite{li2018domain}, domain adversarial learning \cite{ganin2015unsupervised,ganin2016domain}, to causality-inspired methods \cite{rojas2018invariant, sun2021recovering, christiansen2021causal}.
Recent studies suggest limited improvement in DG methods over empirical risk minimization (ERM) \cite{gulrajani2020search,wiles2022a}, indicating a need for innovation.
While visual foundation models are employed in modern DG techniques \cite{zhang2022delving,bommasani2021opportunities,han2023well}, they often necessitate extensive model selection \cite{li2022sparse, rame2023model}.
Various protocols have been proposed for adaptation, such as linear probing \cite{Wu2020Understanding,du2021fewshot}, LP-FT (linear probing and then full fine-tuning) \cite{kumar2022fine}, and gradual unfreeze (fine-tuning last 
$k$ layers) \cite{lee2022surgical,he2022masked}.
However, conventional DG methods typically rely on a universal parameter set and my blend domain-specific features, and thus hinder generalization.
In contrast, our approach focuses on enriching the model with domain-level and task-specific characteristics to enhance separation of invariant features.

\subsection{Visual Prompt Tuning}
\label{160281459741}
Prompt tuning acts as a powerful tool that furnishes transformer-based large models with the necessary semantic context for particular tasks~\cite{Li2021PrefixTuningOC,Liu2021PTuningVP}.
Such a principle has successfully transitioned to the visual realm, culminating in visual prompt tuning (VPT)~\cite{jia2022visual}. 
Traditional visual prompt methods~\cite{jia2022visual,chen2023understanding} typically implement visual prompts using parameter placeholders, where a single set of prompts is shared across various tasks.
While effective for specific downstream tasks~\cite{Sohn2022VisualPT,Das2023LearningEP}, they are not specifically tailored for OoD scenarios, potentially hindering their ability to bridge distribution shifts within the DG context.  
Interestingly, the potential benefits of more customized prompts have been observed in other areas, such as the use of class-specific prompts for certain fine-grained categories \cite{zhou2022learning}.
This observation highlights the potential of adapting visual prompts to embody domain-level and task-specific details within DG.
Recent works, namely DoPrompt~\cite{zheng2022prompt} and CSVPT~\cite{li2022learning}, have begun to explore the application of visual prompting within DG paradigms.
However, our HCVP presents distinct advancements in this area.
Unlike DoPrompt \cite{zheng2022prompt}, which uses prompts shared across each domain and leads to challenges in prompt selection or fusion, HCVP utilizes generative prompts that encapsulate more nuanced characteristics.
Besides, HCVP eliminates the need for the extra adapters required in DoPrompt.
Moreover, contrasting with CSVPT \cite{li2022learning}, which also adopts generative prompting but lacks additional regularization mechanisms, HCVP introduces an explicit model structure and specialized loss functions. 
These are meticulously designed to guide the visual prompt learning process, ensuring a more targeted and effective integration of domain-specific and task-specific information within the DG framework.
In domain adaptation (DA) area, DAPrompt~\cite{ge2023domain} utilizes textual prompt information which requires manual design while our HCVP introduces an innovative approach by generating visual prompts in an end-to-end manner.

\begin{figure*}[!t]
\centering
\includegraphics[width=0.96\linewidth,height=0.39\textheight,keepaspectratio]{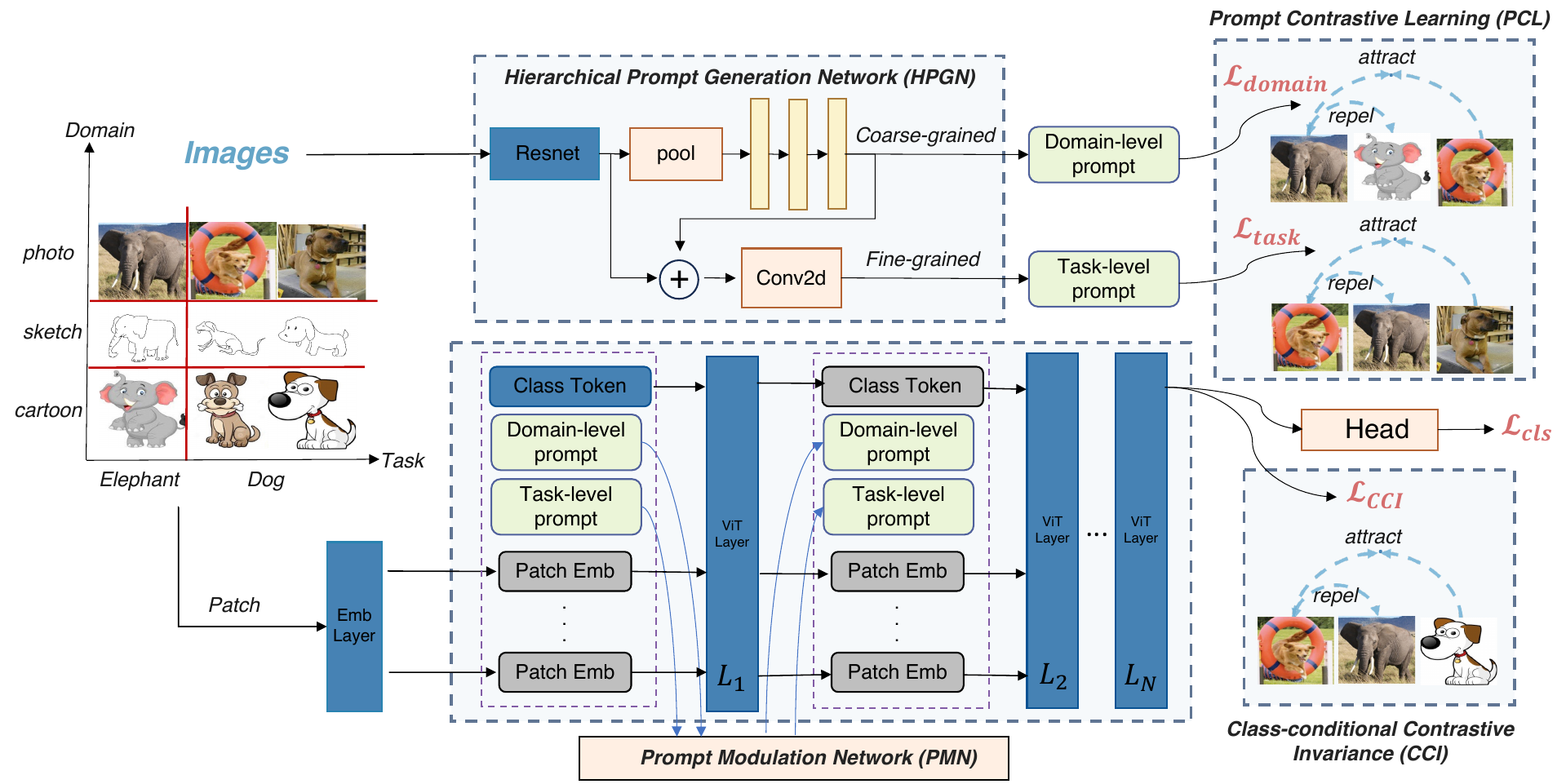}
\caption{
The architecture overview of the proposed Hierarchical Contrastive Visual Prompt (HCVP) model.
HCVP comprises two key components: the Hierarchical Prompt Generation Network (HPGN) and the Prompt Modulation Network (PMN).
The HPGN first uses a pretrained encoder to extract feature maps. These feature maps are then processed through a dual-level generation module for domain-level and task-specific prompt generation, respectively.
The PMN serves as a conduit, integrating the prompts generated by the HPGN into the ViT layers.
Additionally, HCVP incorporates two contrastive learning strategies: Prompt Contrastive Learning (PCL), which optimizes the generation of both domain and task-specific visual prompts, and Class-conditional Contrastive Invariance (CCI), which enhances the model's class-specific discriminative power.
}
\label{351324574768}
\end{figure*}

\section{Preliminaries and Problem Formulation}
In this section, we establish the foundational notations and define the objective of our study, particularly focusing on the role of generative visual prompts in DG.
We also introduce the concept of mutual information as a key metric in our approach.

\subsection{Notations}
To establish a clear understanding of our approach, we define several critical variables that are central to our methodology: $Y$ (the class label we aim to predict), $D$ (the domain index), $Z$ (the learned invariant features crucial for domain generalization), 
$X$ (the raw input features), and $P$ (the learned prompts representing specific characteristics extracted from $X$).
Following~\cite{wang2022generalizing,zhou2022domain}, we denote the input feature space as $\mathcal{X}$ and the target label space as $\mathcal{Y}$. 
A domain comprises data from a joint distribution $P_{XY}$ in the space $\mathcal{X} \times \mathcal{Y}$.
For a given domain $P_{XY}$, $P_X$, $P_{Y\vert X}$, and $P_{X\vert Y}$ represent the marginal, posterior, and class-conditional distributions, respectively. 
We have access to $M$ source domains $\mathcal{S}_{train} = \{ \mathcal{S}^i \vert i=1, \cdots, M\}$, each $\mathcal{S}^i = \{(\bm{x}_{j}^i, y_j^i)\}_{j=1}^{n_i}$, where $\bm{x}$ and $y$ denote the vectors of input features and class labels. 
The joint distributions between any two domains are not equal, i.e., $P_{XY}^{i} \neq P_{XY}^j, 1 \leq i \neq j \leq M$. 
The objective in DG is to learn a function $f: \mathcal{X} \rightarrow \mathcal{Y}$ using the $M$ training domains to minimize prediction error on a new test domain $\mathcal{T}=\{\bm{x}^\mathcal{T}, y^\mathcal{T}\}$.
In our work, this function $f$ is defined as $f(\bm{x},\bm{p})$, where $\bm{p}$ represents generative visual prompts that are integrated into the model to enhance its learning capability:
\begin{equation}
    \label{743842650580}
    \underset{f}{\text{min}} \mathbb{E}_{(\bm{x},y) \in \mathcal{T}}[l(f(\bm{x}, \bm{p}),y)] 
\end{equation}
This formulation reflects our novel approach of enriching the ViT model with visual prompts $\bm{p}$ generated from $\bm{x}$.

\subsection{Role of Generative Visual Prompts in Mutual Information}
Mutual information is a measure from information theory that quantifies the amount of information obtained about one random variable through observing another random variable~\cite{tishby2000information,tishby2015deep}.
In the context of DG, we consider two key mutual information terms~\cite{li2022invariant}: 
\begin{itemize}
    \item $I(Z,Y)$ quantifies how much information the learned representation $Z$ shares with the class label $Y$. A higher value indicates that $Z$ is more informative about $Y$.
    \item $I(Z, D \vert Y)$ measures the information that $Z$ contains about the domain $D$, given the label $Y$.
    In DG, we aim to reduce this term, indicating that the learned representation is less influenced by domain-specific features and more by features that generalize across domains.
\end{itemize}
We hypothesize that the incorporation of visual prompts $P$ into the learning process will lead to a reduction of $I(Z,D\vert Y)$, indicating a more invariant representation.
This hypothesis is grounded on the intuition that $P$, by providing additional context, assists the model in better isolating specific variations from invariant features.
Formally, we aim to show:
\begin{equation}
\label{585756139562}
I(Z^{'}, D \vert Y) < I(Z, D \vert Y)
\end{equation}
Here, $Z^{'}$ is the representation learned with the inclusion of $P$. 
Our subsequent analysis and empirical results aim to support this hypothesis.
Furthermore, our model architecture and objective functions, detailed in the following methodology section, are explicitly designed to align with such two mutual information terms.

\section{Methodology}
Our proposed HCVP methodology, as depicted in Figure~\ref{351324574768}, illustrates the comprehensive architecture tailored for DG. 
Specifically, the hierarchical prompt generation network and prompt modulation network, complemented with prompt contrastive learning, are designed to ensure that the generated prompt vectors align precisely with specific characteristics.
The cross-entropy loss ($\mathcal{L}_{cls}$) directly supports the mutual information term $I(Z,Y)$, while the class-conditioned contrastive invariance loss ($\mathcal{L}_{CCI}$) is designed to align with $I(Z,D \vert Y)$ respectively.
This alignment with the mutual information framework is instrumental in achieving our goal of enhancing DG capabilities of the model.

\subsection{Hierarchical Prompt Generation Network}
\label{691578552537}

As discussed in Sec.~\ref{160281459741}, although traditional visual prompt methods~\cite{jia2022visual} are effective for specific downstream tasks, they are not specifically tailored for OoD scenarios.
While merely replicating visual prompts for individual domain can allow prompts to mirror domain-level information, prompt selection or fusion becomes challenging in DG settings where the target domain is unknown or inaccessible~\cite{zheng2022prompt}.
To tackle these challenges, our Hierarchical Prompt Generation Network (HPGN) offers a sophisticated solution.
While CSVPT~\cite{li2022learning} also utilizes generative visual prompts, HPGN, goes a step further by integrating prompt contrastive learning.
In our work, the model structure and loss functions are specifically designed to generate visual prompts that adeptly encapsulate domain-level and task-specific information. 

\subsubsection{Domain-level Prompt Generation} In the first level of our hierarchical model, we design a mechanism to generate domain-level prompts.
These prompts are aimed at encapsulating coarse-grained and high-level features that are common across a particular domain. 
The process begins by taking input features, denoted as $\bm{x}$, and passing them through a pre-trained and frozen ResNet encoder \cite{7780459}, resulting in the output feature maps represented as $F = R(\bm{x})$\footnote{The domain index is omitted for simplicity.}.
To further distill these feature maps, we append a global average pooling (GAP) layer, denoted as $G(\cdot)$, followed by a fully connected (FC) layer, which is a Multi-Layer Perceptron (MLP) and is denoted as \(FC(\cdot)\). 
The GAP layer functions to reduce the spatial dimensions of the feature maps $F$, effectively encapsulating the entire spatial content into a single value per feature map.
This operation retains the dominant global features across the spatial domain. 
The subsequent MLP is capable of learning non-linear combinations of these global features, thus allowing for the capture of intricate global patterns in the data. 
The resulting domain prompt vector for an instance \(\bm{x}\) is:
\begin{equation}
\label{656501625529}
    C(\bm{x}) = FC(G(F))
\end{equation}
where \(F = R(\bm{x})\). This structure captures the overall domain characteristics, setting the stage for further refinement in the subsequent hierarchical level.

\subsubsection{Task-specific Prompt Generation}
The second level of the hierarchy focuses on creating task-specific prompts to capture task-specific features. 
By appending a set of convolutional layers, denoted as $\Phi(\cdot)$, to the ResNet model, we facilitate the extraction of localized and intricate patterns from the feature maps.
In the context of DG, where each classification task operates within a unique domain, it becomes pertinent to refine task-specific prompts by integrating the domain-level prompts as inputs.
This integration forms a hierarchical structure, synergizing the broad domain-level insights with a refined understanding of task-specific characteristics. 
The task-specific prompt for an instance \(\bm{x}\) is formally given by:
\begin{equation}
\label{656501625530}
    P(\bm{x}) = \Phi(F, C(\bm{x}))
\end{equation}
where \(F = R(\bm{x})\) and \(C(\bm{x})\) is the domain-level prompt in Eq. (\ref{656501625529}).
These two prompt vectors are concatenated to form the final prompt vector, represented as $\bm{p} = \text{concat}(C(\bm{x}), P(\bm{x}))$, where $\text{concat}(\cdot)$ denotes the concatenation operation.
The resulting prompt vector is subsequently fed into the PCL module for optimization and combined with image patch embeddings as the input for ViT layers, shown in Figure \ref{351324574768}.

\subsection{Prompt Modulation Network}
Within our architecture, the Prompt Modulation Network (PMN) serves as a conduit to effectively integrate the generated visual prompts into the ViT backbone. 
Unlike traditional visual prompt methods \cite{jia2022visual} that rely on fixed prompt placeholders, our use of dynamically generated prompts necessitates this specially designed integration mechanism. 
Specifically, the PMN receives the two-tier generated visual prompts as input and transforming them through a MLP.
Importantly, the MLP is configured with a number of layers that is identical to the layer count in the ViT.
This transformation process is conducted during training, where the transformed prompts are systematically integrated into each ViT layer $L_i$, to facilitate a nuanced adaptation of the visual content.
The class token vector and patch embeddings after $i$-th ViT layer are represented by $\mathbf{x}_i$ and $\mathbf{E}_i$, and the integration is formalized as:
\begin{align}
     \label{677284504449}
    [\mathbf{x}_i, \_, \_, \mathbf{E}_i] &= L_i([\mathbf{x}_{i-1}, C_{i-1}, P_{i-1}, \mathbf{E}_{i-1}) \\
    \label{110844202476}
    [C_{i}, P_{i}] &= \text{PMN}(C_{i-1}), \text{PMN}(P_{i-1})
\end{align}
This modulation mechanism, drawing on principles from batch normalization \cite{ioffe2015batch} and attention mechanisms \cite{bahdanau2015neural,vaswani2017attention}, allows the PMN to adapt the ViT's internal representations, effectively utilizing the generated visual prompts.
This enhances the model's ability to accommodate both domain-level and task-specific details.

\subsection{Prompt Contrastive Learning}
We introduce a prompt contrastive learning strategy, which acts as the supplementary to HPGN and is tailored to learn both domain and task prompts, thereby guiding the learning process for visual prompts effectively. 

Domain-level prompt contrastive learning focuses on producing similar domain-level prompts for instances within the same domain, while ensuring distinct ones from different domains. 
As a result, the generated domain-level prompts accurately encapsulate the unique characteristics associated with each domain. 
The loss for a positive pair and the total loss are defined as follows:
\begin{equation}
\label{656501625531}
l_{c}^{ij} = -\log \frac{\exp({C(\bm{x}_i) \cdot C(\bm{x}_{j})}/{\tau})}{\sum_{k} \mathbbm{1}_{[k \neq i]}\exp({C(\bm{x}_i) \cdot C(\bm{x}_{k})}/{\tau})} 
\end{equation}

\begin{equation}
\label{656501625532}
\mathcal{L}_{domain} = \frac{1}{n_i}\sum_{i=1}^{n_i} l_{c}^{ij}
\end{equation}
where \(C(\bm{x}_i)\) represents the domain-level prompt for the \(i\)-th instance in the sampled batch, \(j\) denotes a positive sample of \(i\), specifically from the same domain.
The temperature parameter is denoted by \(\tau\) and dot product is utilized as the similarity measure.

Task-specific prompt contrastive learning aims to encourage the model to generate similar task prompts for instances of the same class within a domain, and distinct ones for instances from different classes within a domain. 
The loss for a positive pair is defined as follows:
\begin{equation}
\label{656501625533}
l_{p}^{ij} = -\log \frac{\exp\left({P(\bm{x}_i) \cdot P(\bm{x}_{j})}/{\tau}\right)}{\sum_{k} \mathbbm{1}_{[k \neq i]}\exp({P(\bm{x}_i) \cdot P(\bm{x}_{k})}/{\tau})}
\end{equation}

\begin{equation}
\label{656501625534}
\mathcal{L}_{task} = \frac{1}{n_i}\sum_{i=1}^{n_i} l_{p}^{ij}
\end{equation}
Here, \(P(\bm{x}_i)\) represents the task prompt for the \(i\)-th instance, \(j\) denotes a positive sample of \(i\), specifically from the same class within the same domain.
The overall prompt contrastive learning loss \(\mathcal{L}_{\text{{PCL}}}\) is a combination of \(\mathcal{L}_{\text{{domain}}}\) and \(\mathcal{L}_{\text{{task}}}\) with equal weight.

\subsection{Class-conditioned Contrastive Invariance}
Our approach introduces Class-conditioned Contrastive Invariance (CCI) as a key component to augment the discriminative ability between different classes while promoting invariance to domain-specific variations.
CCI focuses on promoting invariance within the same class, regardless of the domain they originate from, while maintaining sensitivity to differences between distinct classes. 
The CCI loss function is designed to encourage representations to be domain-invariant in the context of class labels, thereby decreasing $I(Z,D\vert Y)$.

Given \(\mathbf{x}_N\) as the class token embedding after the last ViT layer, and \(y\) as the class label of the instance, the CCI loss can be defined as:
\begin{equation}
\label{656501625535}
\mathcal{L}_{\text{CCI}} = -\mathbb{E}\left[\log \frac{\exp(\mathbf{x}_N \cdot \mathbf{x}_{N'} / \tau)}{\sum_{k \neq N}\exp(\mathbf{x}_N \cdot \mathbf{x}_{k} / \tau)}\right]
\end{equation}
where \(\mathbf{x}_{N'}\) represents a positive sample of \(\mathbf{x}_N\) (same class regardless of the domains), and \(\mathbf{x}_{k}\) signifies a negative sample of \(\mathbf{x}_N\) (different class).
Through this formulation, CCI effectively reinforces the desired domain invariance while maintaining robust class discriminability within the learned representations.

\subsection{End-to-End Training Strategy}

Our training strategy combines multiple loss components: prompt contrastive learning loss $\mathcal{L}_{\text{PCL}}$, class-conditioned contrastive invariance loss $\mathcal{L}_{\text{CCI}}$, and classification loss $\mathcal{L}_{\text{cls}}$, as shown in Algorithm~\ref{alg:HCVP}. 
The classification loss is defined using cross-entropy between the output from the classification head applied to $\mathbf{x}_N$ and the ground-truth label $y$, aiming to increase the mutual information term $I(Z, Y)$:
\begin{equation}
\label{656501625537}
\mathcal{L}_{\text{cls}} = -\sum_{i=1}^{C} y_i \log(\text{Head}(\mathbf{x}_N)_i)
\end{equation}

We then formulate the total loss $\mathcal{L}_{\text{total}}$:
\begin{equation}
\label{656501625536}
\mathcal{L}_{\text{total}} = \mathcal{L}_{\text{cls}} + \lambda_{PCL} \mathcal{L}_{\text{PCL}} + \lambda_{CCI} \mathcal{L}_{\text{CCI}}
\end{equation}
Here, $\lambda_{PCL}$ and $\lambda_{CCI}$ are balancing coefficients for each loss component respectively.
These coefficients are determined through a systematic grid search.

\begin{algorithm}
\caption{Hierarchical Contrastive Visual Prompt (HCVP) Algorithm}
\label{alg:HCVP}
\begin{algorithmic}[1] 
\Require Number of steps $T$, pretrained ResNet encoder, Vision Transformer (ViT) model $f$, classifier $\phi$, and hyper-parameters $\lambda_{PCL}$, $\lambda_{CCI}$.
\State Initialize Hierarchical Prompt Generation Network (HPGN) and Prompt Modulation Network (PMN) parameters.
\For{$t \gets 1$ \textbf{to} $T$}
    \State Sample a batch of images and labels from $M$ source domains $\{\mathcal{S}_1, \ldots, \mathcal{S}_M\}$:
    \For{$i \gets 1$ \textbf{to} $M$}
        \State $\{(\bm{x}_j^{i}, y_j^{i})\}_{j=1}^{n_i} \gets$ Sample from domain $\mathcal{S}_i$
    \EndFor
    \State Generate domain and task prompts using HPGN based on Eq. (\ref{656501625529}) and Eq. (\ref{656501625530}).
    \State Modulate prompts with PMN and integrate into ViT layers as per Eq. (\ref{677284504449}) and Eq. (\ref{110844202476}).
    \State Calculate Prompt Contrastive Learning (PCL) loss $\mathcal{L}_{PCL}$ using Eq. (\ref{656501625532}) and Eq. (\ref{656501625534}).
    \State Compute Class-Conditional Contrastive Invariance (CCI) loss $\mathcal{L}_{CCI}$ using Eq. (\ref{656501625535}).
    \State Compute classification loss $\mathcal{L}_{cls}$ using Eq. (\ref{656501625537}).
    \State Total loss $\mathcal{L}_{\text{total}} \gets \mathcal{L}_{\text{cls}} + \lambda_{PCL} \mathcal{L}_{\text{PCL}} + \lambda_{CCI} \mathcal{L}_{\text{CCI}}$.
    \State Backward $\mathcal{L}_{\text{total}}$ and update the parameters.
\EndFor
\State \Return Trained HCVP model and results based on the model selection criteria using Training-domain validation as per \cite{gulrajani2020search}.
\end{algorithmic}
\end{algorithm}

\section{Experiments}

\begin{table*}[!t]
\centering
\caption{Performance comparison of HCVP with existing DG algorithms and tuning protocols on five DG datasets. 
The best and second-best results are marked in \textbf{\color{red}Red} and \textbf{\color{blue}Blue} respectively. OH and TI represent OfficeHome and TerraIncognita datasets. Algorithms are ordered according to their publication years for ease of reference.}
\label{017557960858}
\begin{tabular}{l | c c c c c | r}
\toprule
Methods & {PACS} & {VLCS} & OfficeHome & TerraIncognita & CelebA & Average\\
\midrule
&\multicolumn{6}{>{\columncolor{LightBlue}}c}{\textit{DG Algorithms}} \\
ERM (1998)~\cite{vapnik1998statistical} & 89.3 & 80.0 & 82.5 & 54.7 & 84.6 & 78.2\\
DANN (2016)~\cite{ganin2016domain} & 84.2 & 79.0 & 80.1 & 47.8 & 85.5 & 75.3\\
MLDG (2018)~\cite{li2018learning} & 89.9 & 79.8 & 82.5 & 52.9 & 85.0 & 78.0\\
MMD (2018)~\cite{li2018domain} & 88.4 & 79.9 & 82.1 & 53.0 & \textbf{\color{blue}85.7} & 77.8\\
CDANN (2018)~\cite{li2018deep} & 84.3 & 78.4 & 80.4 & 49.4 & 84.5 & 75.4\\
IRM (2019)~\cite{arjovsky2019invariant} & 82.9 & 78.1 & 73.6 & 38.9 & 82.1 & 71.1\\
SagNet (2019)~\cite{nam2019reducing} & 83.5 & 80.1 & 77.6 & 48.3 & 83.8 & 74.7\\
RSC (2020)~\cite{huang2020self} & 89.8 & 79.9 & 82.5 & \textbf{\color{red}55.5} & 85.0 & 78.5\\
VREx (2021)~\cite{krueger2021out} & 87.3 & 79.4 & 80.3 & 53.1 & 84.5 & 76.9\\
MTL (2021)~\cite{blanchard2021domain} & 87.2 & 80.1 & 80.7 & 55.0 & 84.5 & 77.5\\
IB\_ERM (2021)~\cite{ahuja2021invariance} & 82.2 & 78.5 & 71.6 & 33.6 & 83.4 & 69.9\\
IB\_IRM (2021)~\cite{ahuja2021invariance} & 81.4 & 78.3 & 58.4 & 37.7 & 82.6 & 67.7\\
SWAD (2021)~\cite{cha2021swad} & 90.1 & 79.3 & 79.7 & 53.0 & 84.0 & 77.2\\
 Transfer (2021)~\cite{zhang2021quantifying} & 83.0 & 78.8 & 75.8 & 48.8  & 80.2 & 73.3 \\
MIRO (2022)~\cite{cha2022domain} & 83.6 & 77.4 & 66.0 & 47.7 & 81.4 & 71.2\\
 EQRM (2022)~\cite{eastwood2022probable} & 90.0 & 78.7 & 82.8 & 54.0 & 83.7 & 77.8 \\
 CausIRL-MMD (2022)~\cite{chevalley2022invariant} &  89.0 & 80.8 & 82.5 & 53.1 & 84.0 & 77.9 \\
POEM (2023)~\cite{jo2023poem} & 84.3 & 76.1 & 73.3 & 39.3 & 83.7 & 71.3\\
 SAGM (2023)~\cite{wang2023sharpness} & \textbf{\color{red}90.4} & 80.3 & 82.5 & 54.8 & 82.7 & 78.2\\
\midrule
& \multicolumn{6}{>{\columncolor{LightOrange}}c}{\textit{Tuning Protocols}} \\
Full (1998)~\cite{vapnik1998statistical} & 89.3 & 80.0 & 82.5 & 54.7 & 84.6 & 78.2\\
Linear Probing (2022)~\cite{kumar2022fine} & 71.6 & 76.9 & 75.6 & 35.2 & 71.0 & 66.1\\
Partial\_1 (2022) ~\cite{lee2022surgical} & 74.5 & 79.4 & 76.1 & 36.7 & 79.7 & 69.3\\
Partial\_2 (2022) ~\cite{lee2022surgical} & 76.7 & 79.9 & 76.8 & 38.4 & 81.1 & 70.6\\
Partial\_4 (2022) ~\cite{lee2022surgical} & 80.5 & \textbf{\color{blue}80.3} & 77.7 & 38.8 & 84.6 & 72.4\\
DoPrompt (2022)~\cite{zheng2022prompt} & 89.8 & 80.3 & \textbf{\color{red}82.7} & 54.5 & 85.3 & \textbf{\color{blue}78.5}\\
\midrule
& \multicolumn{6}{>{\columncolor{LightRed}}c}{\textit{Our Method}} \\
HCVP & \textbf{\color{blue}90.2} & \textbf{\color{red}81.1} & \textbf{\color{blue}82.5} & \textbf{\color{blue}55.1} & \textbf{\color{red}86.6} & \textbf{\color{red}79.1}\\
\bottomrule
\end{tabular}
\end{table*}

\subsection{Experimental Setup}

\subsubsection{Datasets} We evaluate on five benchmark datasets in DG~\cite{gulrajani2020search,ye2022ood}. 
\textbf{PACS}~\cite{li2017deeper}: encapsulates four distinctive visual domains: Art Paintings, Cartoon, Photos, and Sketches; contains 9,991 instances across 7 classes, each being a (3, 224, 224)-dimensional image.
\textbf{VLCS}~\cite{fang2013unbiased}: an amalgamation of four photographic domains: Caltech-101, LabelMe, SUN09, and VOC2007; includes 10,729 instances in five classes, each a (3, 224, 224)-dimensional image.  
\textbf{Office-Home}~\cite{venkateswara2017deep}: comprises four domains: Art, Clipart, Product, and Real-World Images, with 15,588 images in 65 classes, each of dimension (3, 224, 224).
\textbf{TerraIncognita}~\cite{beery2018recognition}: features wildlife images from four camera trap locations (L100, L38, L43, L46), each representing a distinct domain; has 24,788 (3, 224, 224)-dimensional images across 10 classes.
\textbf{CelebA}~\cite{liu2015deep}: focuses on ``hair color'' classification against ``gender'' as a spurious attribute~\cite{sagawa2019distributionally,ye2022ood}. Our subset has 27,040 images in three environments, akin to the Colored MNIST dataset's correlation shift~\cite{arjovsky2019invariant,ye2022ood}.

\subsubsection{Baselines} We compare HCVP with two categories of DG methods: 
\textbf{DG Algorithms} include ERM~\cite{vapnik1998statistical}, IRM~\cite{arjovsky2019invariant}, MLDG~\cite{li2018learning}, MMD~\cite{li2018domain}, DANN~\cite{ganin2016domain}, CDANN~\cite{li2018deep}, VREx~\cite{krueger2021out}, MTL~\cite{blanchard2021domain}, SagNet~\cite{nam2019reducing}, RSC~\cite{huang2020self}, IB\_ERM, IB\_IRM~\cite{ahuja2021invariance}, SWAD~\cite{cha2021swad}, MIRO~\cite{cha2022domain} and POEM~\cite{jo2023poem}; 
\textbf{Tuning Protocols} include full fine-tuning, linear probing (LP)---a frozen backbone with a tuned linear head~\cite{kumar2022fine}, partial-k (fine-tuning last \textit{k} layers).
We also acknowledge related visual prompt methods like DoPrompt~\cite{zheng2022prompt}. 
A direct comparison with CSVPT~\cite{li2022learning} is not feasible, as its implementation is not open-sourced.

\subsubsection{Implementation Details} We adopt unified settings for all methods: ViT-B/16 backbone\footnote{See `vit\_base\_patch16\_224' in \url{https://github.com/huggingface/pytorch-image-models/blob/main/timm/models/vision\_transformer.py}} pre-trained on ImageNet-1k and finetuned on ImageNet-21k~\cite{dosovitskiy2021an, deng2009imagenet}; AdamW optimizer \cite{loshchilov2017decoupled} with learning rate 1$e$-5 and weight decay 0.01; 80\%:20\% train-validation split, batch size of 32, 3000 steps per trial; averages reported over 3 runs with different seeds. 
Training-domain validation strategy is used for model selection.
We use PyTorch to implement all experiments on NVIDIA A100-40GB and A5000-24GB GPUs.
Specifically, the PMN comprises a series of linear layers mirroring the ViT's architecture, with each layer autonomously modulating domain and task prompts for precise alignment with its corresponding ViT layer.

\subsection{Main Results}

In our evaluation, HCVP is benchmarked against established DG algorithms and various tuning protocols.
The comparative results are presented in Table \ref{017557960858}.
For a more detailed analysis, particularly focusing on performance across each unseen domain, refer to Tables \ref{463767169933}, \ref{685752508887}, \ref{232161974285}, and \ref{600458111874}.
Our comparison spans two primary types of distribution shifts: diversity shift observed in the PACS, VLCS, OfficeHome and TerraIncognita datasets, and the spurious shift in the CelebA dataset \cite{ye2022ood}.

\begin{table}[!htb]
\centering
\caption{Performance comparison on PACS dataset. 
The average accuracy across all domains is reported.
``$\rightarrow$'' denotes the unseen domain.
}
\label{463767169933}
\setlength{\tabcolsep}{2.7pt}
\begin{tabular}{l|cccc|r}
\toprule
Methods & \(\rightarrow\) Art & \(\rightarrow\) Cartoon & \(\rightarrow\) Photo & \(\rightarrow\) Sketch & Avg\\
\midrule

&\multicolumn{5}{>{\columncolor{LightBlue}}c}{\textit{DG Algorithms}} \\
ERM & 94.85 & 87.56 & 99.68 & 75.13 & 89.31 \\
DANN & 89.20 & 81.79 & 99.00 & 66.60 & 84.15\\
MLDG & 95.08 & 87.76 & 99.38 & 77.30 & 89.88\\
MMD & 93.35 & 86.55 & 99.70 & 73.91 & 88.38\\
CDANN & 89.73 & 81.59 & 98.80 & 66.96 & 84.27\\
IRM & 91.11 & 76.69 & 99.35 & 64.33 & 82.87 \\
SagNet&90.10&80.95&99.48&63.50&83.51\\
RSC&95.28&87.42&99.68&76.83&89.80\\
VREx & 92.68 & 84.15 & 99.45 & 72.81&87.27\\
MTL&92.48&84.90&99.55&71.68&87.15\\
IB\_ERM&89.69&78.00&99.43&61.58&82.18\\
IB\_IRM&90.10&78.27&98.73&58.51&81.40\\
SWAD&93.23&85.93&99.18&82.03&90.10 \\
Transfer & 91.21 & 82.62 & 98.88 & 59.38 & 83.02 \\
MIRO&83.41&78.95&93.19&78.91&83.61\\
EQRM & 93.35 & 86.83 & 99.78 & 79.87 & 89.96\\
CausIRL-MMD & 94.20 & 85.07 & 99.85 & 77.00 & 89.03 \\
POEM &86.21&79.74&97.16&74.17&84.32 \\
SAGM & 93.53 & 87.95 & 98.73 & 81.39 & \textbf{\color{red}90.40}\\

\midrule
& \multicolumn{5}{>{\columncolor{LightOrange}}c}{\textit{Tuning Protocols}} \\
Full & 94.85 & 87.56 & 99.68 & 75.13 & 89.31\\
Linear Probing&84.75&68.85&92.54&40.41&71.64\\
Partial\_1&87.80&70.52&98.03&41.79&74.54 \\
Partial\_2& 89.77&73.93&98.30&44.88&76.72\\
Partial\_4 &91.64&77.90&99.00&53.41&80.49 \\
DoPrompt&95.04&86.35&99.63&78.20&89.81 \\
\midrule
& \multicolumn{5}{>{\columncolor{LightRed}}c}{\textit{Our Method}} \\
HCVP&93.17&86.89&99.33&81.30&\textbf{\color{blue}90.17}\\
\bottomrule
\end{tabular}

\label{162960768294}
\end{table}

\begin{table}[!htb]
\centering
\caption{Performance comparison on VLCS dataset. 
The average accuracy across all domains is reported.
``$\rightarrow$'' denotes the unseen domain.}
\label{685752508887}
\setlength{\tabcolsep}{0.6pt}
\begin{tabular}{l| c c c c | r}
\toprule
Methods & $\rightarrow$ Caltech101 &$\rightarrow$ LabelMe &$\rightarrow$ SUN09 &$\rightarrow$ VOC2007 & Avg\\
\midrule

&\multicolumn{5}{>{\columncolor{LightBlue}}c}{\textit{DG Algorithms}} \\
ERM&96.50&65.54&78.32&79.76&80.03 \\
DANN & 94.05&65.68&76.96&79.33&79.01 \\
MLDG & 96.82&64.52&78.47&79.55&79.84\\
MMD & 96.70&66.37&78.23&78.37&79.92 \\
CDANN & 95.17&65.05&75.97&77.32&78.38\\
IRM&97.67&63.81&73.86&77.11&78.11\\
SagNet & 97.44&64.96&77.99&80.16&80.14\\
RSC & 96.41&65.27&78.10&79.93&79.93 \\
VREx & 96.55&65.40&76.29&79.38&79.41\\
MTL & 96.76&65.43&78.93&79.29&80.10\\
IB\_ERM & 97.59&62.76&75.41&78.22&78.50 \\
IB\_IRM & 98.47&64.83&74.51&75.27&78.27\\
SWAD & 98.49&63.86&75.40&79.49&79.31 \\
Transfer & 98.85 & 65.60 & 74.37 & 76.23 & 78.76 \\
MIRO & 98.85&63.44&69.31&78.08&77.42 \\
EQRM & 94.88 & 64.33 & 77.38 & 78.38 & 78.74\\
CausIRL/MMD & 97.70 & 66.82 & 78.41 & 80.41 & 80.84 \\
POEM & 96.91&58.40&73.31&75.60&76.05\\
SAGM & 99.21 & 62.92 & 76.12 & 83.04 & 80.32\\
\midrule
& \multicolumn{5}{>{\columncolor{LightOrange}}c}{\textit{Tuning Protocols}} \\
Full &96.50&65.54&78.32&79.76&80.03 \\
Linear Probing & 98.70&65.33&71.71&71.79&76.88\\
Partial\_1 & 98.97&65.10&76.25&77.26&79.40 \\
Partial\_2 & 98.41&64.44&77.55&79.07&79.87\\
Partial\_4 & 98.03&64.44&78.26&80.33&\textbf{\color{blue}80.27}\\
DoPrompt & 96.70&66.53&78.28&79.39&80.23 \\
\midrule
& \multicolumn{5}{>{\columncolor{LightRed}}c}{\textit{Our Method}} \\
HCVP & 96.32&66.26&80.08&81.65&\textbf{\color{red}81.08} \\
\bottomrule
\end{tabular}

\label{293272028507}
\end{table}

\begin{table}[!htb]
\centering
\caption{Performance comparison on OfficeHome dataset. 
The average accuracy across all domains is reported.
``$\rightarrow$'' denotes the unseen domain.}
\setlength{\tabcolsep}{2.3pt}
\label{232161974285}
\begin{tabular}{l| c c c c | r}
\toprule
Methods & $\rightarrow$ Art &$\rightarrow$ Clipart &$\rightarrow$ Product &$\rightarrow$ Real & Avg\\
\midrule

&\multicolumn{5}{>{\columncolor{LightBlue}}c}{\textit{DG Algorithms}} \\
ERM & 81.38&69.78&88.72&90.24&82.53 \\
DANN & 78.39&66.28&87.09&88.80&80.14 \\
MLDG & 80.71&70.56&88.25&90.50&82.51 \\
MMD & 80.45&68.92&88.96&90.19&82.13 \\
CDANN & 78.77&65.85&87.63&89.22&80.37 \\
IRM & 71.88&58.39&80.97&82.99&73.56 \\
SagNet & 74.34&62.55&85.64&88.03&77.64\\
RSC & 80.50&70.12&89.02&90.45&82.52\\
VREx & 79.09&65.56&87.33&89.36&80.34\\
MTL & 79.71&66.51&87.37&89.18&80.69 \\
IB\_ERM & 69.91&56.55&78.46&81.47&71.60 \\
IB\_IRM & 56.56&47.22&61.42&68.23&58.36 \\
SWAD & 76.26&68.87&86.74&87.03&79.73 \\
Transfer & 72.86 & 62.49 & 82.09 & 85.80 & 75.81 \\
MIRO & 61.48&50.03&75.56&76.91&66.00\\
EQRM & 81.31 & 70.50 & 88.91 & 90.48 & 82.80\\
CausIRL/MMD & 80.43 & 70.68 & 88.65 & 90.19 & 82.49 \\
POEM & 70.13&59.88&80.41&82.59&73.25\\
SAGM & 80.74 & 70.99 & 88.99 & 89.24 & 82.49\\
\midrule
& \multicolumn{5}{>{\columncolor{LightOrange}}c}{\textit{Tuning Protocols}} \\
Full & 81.38&69.78&88.72&90.24&82.53 \\
Linear Probing & 75.49&56.04&83.59&87.45&75.64 \\
Partial\_1 & 75.58&56.55&83.92&88.33&76.10 \\
Partial\_2 & 76.35&58.33&84.37&88.20&76.81 \\
Partial\_4 & 77.14&60.20&84.88&88.42&77.66 \\
DoPrompt & 80.95&70.88&88.94&90.10&\textbf{\color{red}82.72}\\
\midrule
& \multicolumn{5}{>{\columncolor{LightRed}}c}{\textit{Our Method}} \\
HCVP & 81.77&69.76&88.01&90.62&\textbf{\color{blue}82.54} \\
\bottomrule
\end{tabular}

\label{854623515096}
\end{table}

\subsubsection{DG Algorithms}
As illustrated in Table \ref{017557960858}, our HCVP achieves the highest overall accuracy of 79.1\%, and achieves the best or second-best performance on each dataset. 
Specifically, HCVP attains the best average accuracies of 90.2\%, 81.1\% and 86.6\% on PACS, VLCS and CelebA datasets, respectively.
These results demonstrate that HCVP effectively learns invariant features that are informative for prediction across domains.
In the case of the OfficeHome dataset, our HCVP still achieves a comparable performance of 82.5\%, only 0.2\% lower than the best performance achieved by MLDG.  
Compared with other baseline DG algorithms designed to capture invariant features, such as MMD and DANN for domain-invariant features, and IRM, VREx, CDANN for conditional invariant features, HCVP achieves significant gains, with an average improvement of 3.8\% in overall performance across five datasets. 
This underscores the fact that, although several nuanced constraints are proposed for invariant features, a single set of model parameters inadvertently incorporates domain-specific features, thereby limiting the model's generalization ability. 
In contrast, our HCVP, which augments domain-level and task-specific characteristics to guide the model to learn invariant features, exhibits robust performance across diverse distribution shifts.

\subsubsection{Tuning Protocols} 
We outline tuning protocols for adapting visual foundation models to distribution shifts in downstream tasks. 
As shown in Table \ref{017557960858}, full fine-tuning, akin to ERM, yields a robust average accuracy of 78.2\%.
In contrast, less computationally intensive methods such as linear probing and gradual unfreezing (Partial\_1, Partial\_2, Partial\_4) achieve relatively lower average accuracies, almost all falling below 72\%. 
This trend clearly illustrates a correlation between the number of tuned parameters and the average performance under distribution shifts. 
In this context, our HCVP integrates domain and task-specific information into the model, by augmenting visual prompts to the ViT backbone.
This enhancement not only aligns with the observed trend of increased performance with broader parameter tuning but also contributes to the model's robustness.
For example, compared to the best tuning protocol---full fine-tuning---our HCVP outperforms it on all five datasets. 
This advantage becomes particularly pronounced when facing substantial distribution shifts. 
For example, when adapting to the sketch domain on the PACS dataset, denoted by \texttt{"}$\rightarrow$ Sketch \texttt{"}, the performance is 81.30\% compared to 75.13\%, as shown in Table~\ref{463767169933}. 
Similarly, when encountering spurious distribution shifts on the CelebA dataset, the performance is 86.6\% versus 84.6\%.

\subsubsection{Significance of Incremental Gains in DG}
In the field of DG, even marginal performance improvements carry considerable significance. 
This is largely attributed to the inherent complexity and the unpredictable nature of generalizing effectively to unseen domains. 
The previous large-scale evaluations \cite{gulrajani2020search, ye2022ood, wiles2022a} indicate that no single DG algorithm consistently outperforms over ERM across all benchmark datasets.
However, our HCVP method demonstrates this remarkable capability.
Therefore, the incremental gains, such as the 1\% improvement reported by HCVP, should be recognized for their substantial value.

\begin{table}[!tb]
\centering
\caption{Performance comparison on TerraIncognita dataset. 
The average accuracy across all domains is reported.
``$\rightarrow$'' denotes the unseen domain.}
\label{600458111874}
\setlength{\tabcolsep}{3.8pt}
\begin{tabular}{l| c c c c | r}
\toprule
Methods & $\rightarrow$ L100 &$\rightarrow$ L38 &$\rightarrow$ L43 &$\rightarrow$ L46 & Avg\\
\midrule

&\multicolumn{5}{>{\columncolor{LightBlue}}c}{\textit{DG Algorithms}} \\
ERM & 63.01 & 46.28 & 62.15 & 47.50 & 54.74 \\
DANN & 52.31 & 42.52 & 55.67 & 39.77 & 47.57 \\
MLDG & 60.37 & 42.74 & 61.40 & 47.10 & 52.90 \\
MMD & 62.09 & 43.64 & 59.04 & 47.33 & 53.03 \\
CDANN & 56.31 & 39.59 & 56.05 & 45.74 & 49.42 \\
IRM & 36.07 & 43.43 & 39.04 & 36.92 & 38.87 \\
SagNet & 58.42 & 41.94 & 54.28 & 38.64 & 48.32 \\
RSC & 65.70 & 43.78 & 62.50 & 50.18 & \textbf{\color{red}55.54} \\
VREx & 58.19 & 47.03 & 59.48 & 47.50 & 53.05 \\
MTL & 60.01 & 47.12 & 61.24 & 51.80 & 55.04 \\
IB\_ERM & 32.82 & 21.21 & 41.72 & 38.64 & 33.60 \\
IB\_IRM & 38.15 & 35.27 & 40.43 & 36.90 & 37.69 \\
SWAD & 59.29 & 47.84 & 61.97 & 42.75 & 52.96 \\
Transfer & 56.55 & 45.44 & 53.40 & 39.90 & 48.82 \\
MIRO & 61.80 & 46.55 & 54.72 & 27.90 & 47.74 \\
EQRM & 61.72 & 45.77 & 62.94 & 45.61 & 54.01\\
CausIRL/MMD & 56.95 & 41.73 & 63.10 & 50.71 & 53.12 \\
POEM & 48.75 & 27.54 & 53.40 & 27.43 & 39.28 \\
SAGM & 65.91 & 46.32 & 60.58 & 46.27 & 54.77\\
\midrule
& \multicolumn{5}{>{\columncolor{LightOrange}}c}{\textit{Tuning Protocols}} \\
Full & 63.01 & 46.28 & 62.15 & 47.50 & 54.74 \\
Linear Probing & 30.64 & 35.13 & 41.53 & 33.38 & 35.17 \\
Partial\_1 & 35.35 & 31.16 & 43.26 & 36.82 & 36.65 \\
Partial\_2 & 39.15 & 26.36 & 46.00 & 41.96 & 38.37 \\
Partial\_4 & 43.11 & 26.33 & 44.46 & 41.45 & 38.84 \\
DoPrompt & 64.51 & 40.93 & 63.00 & 49.37 & 54.45 \\
\midrule
& \multicolumn{5}{>{\columncolor{LightRed}}c}{\textit{Our Method}} \\
HCVP & 62.06 & 51.84 & 58.91 & 47.61 & \textbf{\color{blue}55.11}  \\
\bottomrule
\end{tabular}

\label{665106748910}
\end{table}

\begin{figure}[!b]
    \centering
    \includegraphics[width=0.96\linewidth]{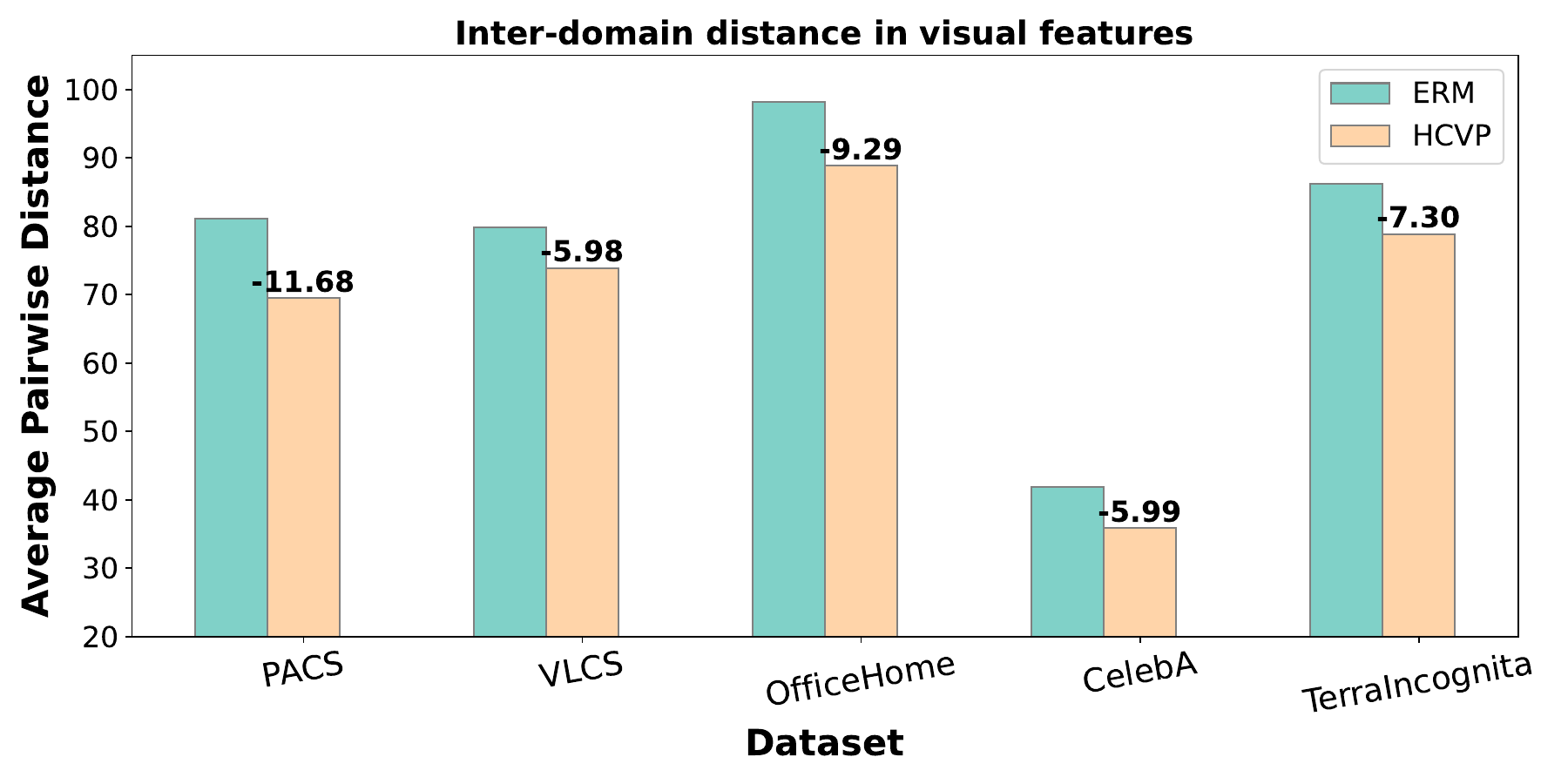}
    \caption{Comparison of inter-domain feature distances for ERM and HCVP across multiple datasets. It illustrates the effectiveness of HCVP in achieving lower inter-domain distances compared to ERM, suggesting a stronger capability for domain-invariant feature learning.}
    \label{262897678632}
\end{figure} 

\subsection{Quantifying Domain-Invariance}

We initially explored the possibility of directly measuring the conditional mutual information 
$I(Z,D \vert Y)$ to assess our model's capacity to learn invariant features~\cite{li2022invariant}.
This measurement is consistent with Eq. \ref{585756139562} to evaluate the hypothesis that the incorporation of $P$ leads to the reduction of $I(Z,D \vert Y)$.
However, given the high-dimensional nature of the latent representation $Z$, we encountered significant computational challenges in reliably estimating this metric. 
To circumvent this issue, we opted to measure inter-domain feature distances as an alternative metric.
This approach is grounded in the theory that lower distances across domains indicate more domain-invariant feature learning~\cite{ben2010theory, ganin2016domain}.
In essence, closer feature representations across domains suggest that the model is not overfitting to domain-specific characteristics and is, instead, capturing more generalizable aspects of the data. 
Specifically, we regard the last domain in each dataset as the unseen domain.
We then train both ERM and HCVP models on this unseen domain and save the iterations of the models that demonstrate the best performance.
Using the saved best model, we calculate the average inter-domain distance by utilizing the latent representations generated by the models.

As shown in Figure \ref{262897678632}, our HCVP consistently achieves smaller inter-domain distances compared to the ERM baseline across all evaluated datasets. 
This observation, together with our main results on the last unseen domain shown in Tables \ref{463767169933}, \ref{685752508887}, \ref{232161974285}, and \ref{600458111874}, strongly indicates that HCVP is effective in extracting domain-invariant features.
While inter-domain distances offer a more indirect measure of domain invariance compared to mutual information, their consistency with our primary results lends feasibility and reliability to this approach.

\subsection{Ablation Study}
In this section, we perfrom ablation study to evaluate the individual contribution of each loss in our HCVP model.
As shown in Table \ref{202678740661}, where ``P, V, O, T, C'' represents PACS, VLCS, OfficeHome, TerraIncognita and CelebA respectively, the complete model \texttt{"}HCVP (full)\texttt{"} achieves the highest accuracy on each dataset and the best agerage accuracy of 79.10.
Removing each loss sequentially leads to \texttt{"}HCVP w/o $\mathcal{L}_{PCL}$\texttt{"} and \texttt{"}HCVP w/o $\mathcal{L}_{CCI}$\texttt{"}, which exhibit slightly reduced accuracies of 78.59 and 78.21, respectively.
This underscores the importance of these losses to the model's performance.
Removing both losses results in \texttt{"}HCVP w/o $\mathcal{L}_{CCI}$ and $\mathcal{L}_{PCL}$\texttt{"}, further reducing the accuracy to 77.69.
Intriguingly, our expanded analysis revealed that removing either $\mathcal{L}_{domain}$ or $\mathcal{L}_{task}$ detrimentally impacts the model's performance more significantly than omitting both in the form of $\mathcal{L}_{PCL}$. 
This finding suggests a complex interplay between $\mathcal{L}_{domain}$ and $\mathcal{L}_{task}$, where their combined presence bolsters the model's domain generalization capability more than their individual contributions might imply.

\begin{table}[tb!]
    \centering
        \caption{
    This ablation study evaluates the individual contributions of various losses in the HCVP model across four datasets. 
    \texttt{"}HCVP (full)\texttt{"}  refers to the complete model.
    We sequentially remove each loss to create \texttt{"}HCVP w/o $\mathcal{L}_{task}$\texttt{"}, \texttt{"}HCVP w/o $\mathcal{L}_{domain}$\texttt{"} \texttt{"}HCVP w/o $\mathcal{L}_{PCL}$\texttt{"} and \texttt{"}HCVP w/o $\mathcal{L}_{CCI}$\texttt{"}. 
    Removing $\mathcal{L}_{PCL}$ and $\mathcal{L}_{CCI}$ results in \texttt{"}HCVP w/o $\mathcal{L}_{CCI}$ and $\mathcal{L}_{PCL}$\texttt{"}, the vanilla HCVP variant with only the classification loss $\mathcal{L}_{cls}$. 
    Values are shown with two decimal place; columns P, V, O, T, C correspond to the PACS, VLCS, OfficeHome, TerraIncognita, and CelebA datasets.
    }
    \setlength{\tabcolsep}{2.9pt} 
    \begin{tabular}{l|c|c|c|c|c|r}
    \toprule
    \multirow{1}{*}{Methods} &  \multicolumn{1}{c|}{\textbf{P}} & \multicolumn{1}{c|}{\textbf{V}} & \multicolumn{1}{c|}{\textbf{O}} & \multicolumn{1}{c|}{\textbf{T}} & \multicolumn{1}{|c|}{\textbf{C}} & \multirow{1}{*}{\textbf{Avg}}\\\midrule
    HCVP w/o $\mathcal{L}_{domain}$ & 88.88 & 80.38 & 81.91 & 54.62 &85.42 & 78.24\\ 
    HCVP w/o $\mathcal{L}_{task}$ & 88.63 & 80.92 & 81.63 & 55.25 & 85.50 & 78.38\\ \midrule
    HCVP w/o $\mathcal{L}_{CCI}$ and $\mathcal{L}_{PCL}$ & 89.43 & 80.30 & 82.16 & 51.05 & 85.50 & 77.69\\ 
    HCVP w/o $\mathcal{L}_{CCI}$ & 89.97 & 79.73 & 82.19 & 53.42 & 85.76 & 78.21  \\ 
    HCVP w/o $\mathcal{L}_{PCL}$ & 89.67 & 80.51 & 82.52 & 54.21 & 86.02 & 78.59 \\ \midrule
    HCVP (full) & \textbf{90.17} & \textbf{81.08} & \textbf{82.54} & \textbf{55.11} & \textbf{86.60} & \textbf{79.10} \\
    \bottomrule
    
    \end{tabular}

    \label{202678740661}
\end{table}

\begin{figure*}[!tb]
\centering
\includegraphics[width=0.99\linewidth,keepaspectratio]{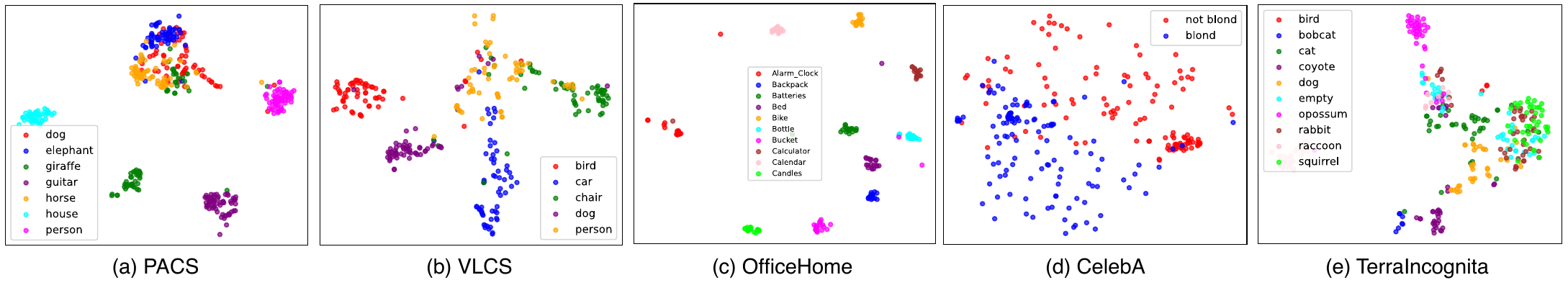}
\caption{The t-SNE visualizations of visual features in our HCVP, for the last unseen domain on PACS, VLCS, OfficeHome, CelebA, and TerraIncognita datasets. Forty instances are sampled within each class. 
Additionally, we select the first ten classes in the OfficeHome dataset.}
\label{653151557414}
\end{figure*}

\subsection{Quanlitative Evaluations}
\begin{figure}[!tb]
    \centering
    \includegraphics[width=0.99\linewidth,keepaspectratio]{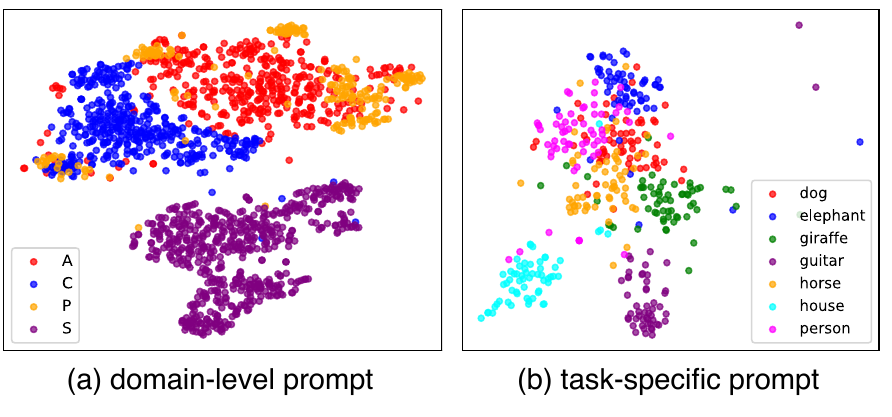}
    \caption{The t-SNE visualizations of domain-level and task-specific prompts on the PACS dataset, representing four domains and seven class labels. Thirty instances are sampled within each class for visualization.}
    \label{307227854026}
\end{figure}

\subsubsection{The t-SNE Visualizations of Visual Features}
Figure \ref{653151557414} illustrates the t-SNE~\cite{Maaten2008VisualizingDU} visualizations of visual features in our HCVP, for the last unseen domain on the PACS, VLCS, OfficeHome, CelebA and TerraIncognita datasets. 
Overall, the visual embeddings of HCVP are clearly clustered in accordance with their class labels across five datasets. 
This clear clustering is indicative of the robustness and effectiveness of our HCVP model in capturing invariant features for task prediction.
In generalizing to the last unseen domain on PACS, namely ``Sketch'', where the test scenario may diverge largely from our pretraining and training features in ``Art, Photo, Cartoon'', an overlap between several classes, such as dogs with elephants and horses, is observed in Figure \ref{653151557414} (a). 
Despite this, our HCVP still shows the best generalization ability, as evidenced in Tables \ref{017557960858} and \ref{463767169933}.

\subsubsection{The t-SNE Visualizations of Visual Prompts}
To investigate the relationship between visual prompts with domain and task specific characteristics, we also leverage t-SNE visualizations, as visual prompts typically lack clear meanings \cite{zhou2022learning,jia2022visual}.
As illustrated in Figure \ref{307227854026}, our visualization study on the PACS dataset reveals: (a) domain prompt vectors corresponding to the same domains are clustered together, these vectors are distinctly separate from other domains, validating that they carry domain-level characteristics; (b) task prompts are similarly clustered by specific task labels. These findings collectively affirm that the visual prompts generated by our method effectively capture domain-level and task-specific characteristics, contributing to HCVP's superior performance in DG tasks.

\subsection{Hyperparameter Tuning}
\begin{figure}[!htb]
    \centering
    \includegraphics[width=0.99\linewidth,keepaspectratio]{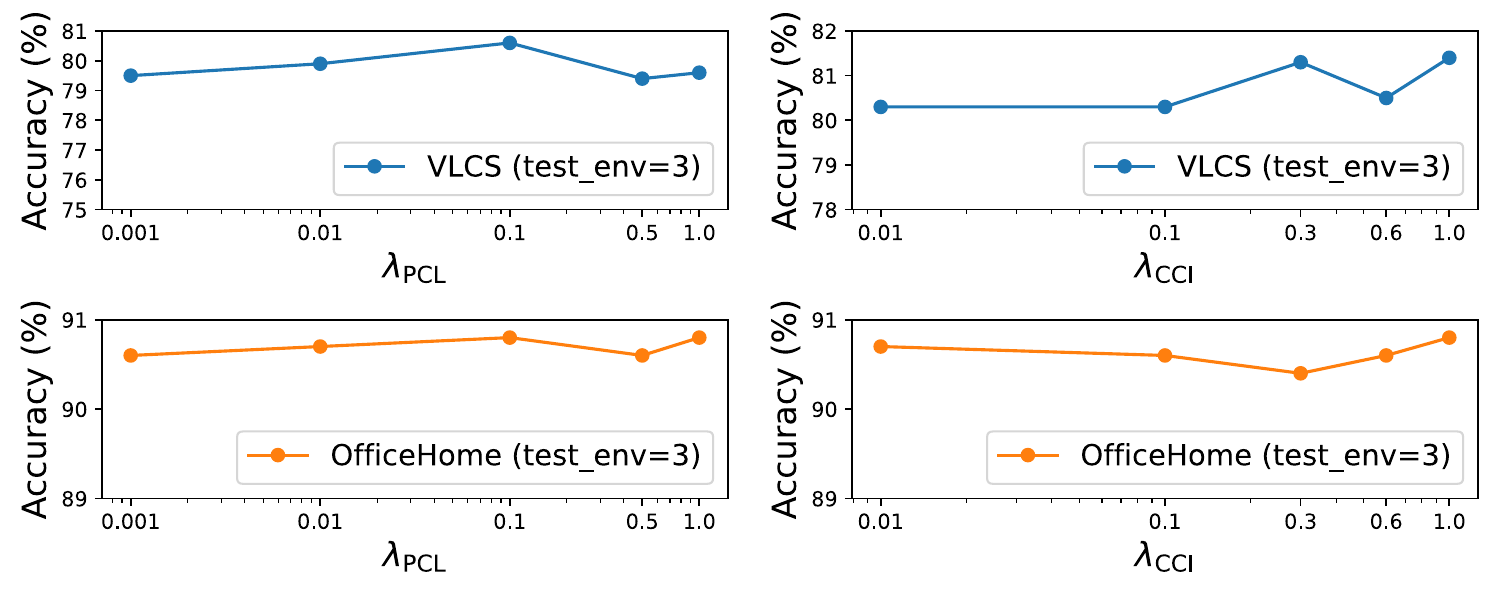}
    \caption{Analysis of the effects of two loss weights, $\lambda_{PCL}$ and $\lambda_{CCI}$, on the VLCS and OfficeHome datasets.}
    \label{345087749500}
\end{figure}

We investigate the effects of two loss weights, $\lambda_{PCL}$ and $\lambda_{CCI}$, on the VLCS and OfficeHome datasets. We explore a wide range of values for both weights, specifically $\{$0.001, 0.01, 0.1, 0.5, 1.0$\}$ for $\lambda_{PCL}$ and $\{$0.01, 0.1, 0.3, 0.6, 1.0$\}$ for $\lambda_{CCI}$. Given that the original value of $\mathcal{L}_{PCL}$ is almost four times larger than $\mathcal{L}_{\text{cls}}$, a large weight for $\lambda_{PCL}$ could impact classification. 
As depicted in Figure \ref{345087749500}, when $\lambda_{PCL}$ is set to 0.1, our HCVP achieves optimal performance. Additionally, we find that 1.0 serves as an effective weight for $\mathcal{L}_{CCI}$.

\subsection{Efficiency Analysis}

The efficiency of DG algorithms is pivotal, balancing computational demands with performance~\cite{gulrajani2020search}. 
Table~\ref{276880594098} presents an efficiency analysis comparing our HCVP model against both traditional and recent state-of-the-art DG methods, evaluating total trainable parameters, gradient parameters within the featurizer, GPU memory usage, step time, and average performance across five datasets.

HCVP demonstrates an increment in total trainable parameters to 90,131,719, indicative of the additional hierarchical prompts and networks engineered to bolster domain generalization. 
Despite this increase, HCVP maintains competitive efficiency with a GPU memory usage of 13.463 GB—comparable to baseline methods with 13.177 GB—and a step time of 1.249 seconds, showcasing notable efficiency, particularly against the 2.404s of DoPrompt and 11.014s of Transfer model.

Crucially, HCVP secures the highest average accuracy of 79.1, outstripping close contenders such as RSC and DoPrompt. 
This underscores the effectiveness of our model's complexity, balanced against a modest rise in resource consumption, to achieve superior domain generalization.
In essence, this analysis confirms the thoughtful trade-off between model complexity and computational efficiency in our HCVP.

\begin{table}[!hb]
\centering
\caption{Efficiency analysis of SOTA and recent DG algorithms. 
``Tot. params" indicates the total number of trainable parameters, ``Grad. in feat." denotes the gradient parameters within the featurizer, ``GPU (GB)" represents the GPU memory usage in gigabytes, ``Step time (s)" specifies the average time per training step in seconds, and ``Avg. Perf." reflects the average accuracy under five datasets.}
\label{276880594098}
\setlength{\tabcolsep}{1.1pt}
\begin{tabular}{l | c c c c c}
\toprule
Methods & {Tot. params} & {Grad. in feat.} & {GPU (GB)} & {Step time (s)} & {Avg. Perf.} \\
\midrule
ERM & 85,804,039 & 85,798,656 & 13.177 & 1.102 & 78.2 \\
MMD & 85,804,039 & 85,798,656 & 12.947 & 1.117 & 77.8 \\
RSC & 85,804,039 & 85,798,656 & 13.179 & 1.170 & 78.5 \\
DoPrompt & 88,212,499 & 85,798,656 & 13.847 & 2.404 & 78.5 \\
SAGM & 85,802,501 & 85,798,656 & 13.177 & 2.001 & 78.2 \\
EQRM & 85,804,039 & 85,798,656 & 11.857 & 1.795 & 77.8 \\
Transfer & 85,809,422 & 85,798,656 & 11.959 & 11.014 & 73.3 \\
CausIRL-MMD & 85,804,039 & 85,798,656 & 11.857  & 0.992 & 77.9 \\
\midrule
HCVP & 90,131,719 & 90,126,336 & 13.463 & 1.249 & 79.1 \\
\bottomrule
\end{tabular}
\end{table}

\section{Conclusion}
In this work, we introduced the Hierarchical Contrastive Visual Prompt (HCVP) methodology, a novel approach to DG that effectively separates invariant features from domain-specific aspects. 
By integrating domain-level and task-specific characteristics through a two-tier hierarchical prompt generation network coupled with prompt contrastive learning, HCVP enhances adaptability to unseen domains. 
Extensive experiments on five benchmark datasets demonstrate the superiority of HCVP over state-of-the-art DG algorithms and adaptation protocols. 
Despite its successes, HCVP may not achieve optimal performance across all unseen domains, owing to the inherent complexity of these domains. 
The method's reliance on extracting solely domain-level and task-specific characteristics from visual information may prove insufficient in certain contexts, thereby inspiring consideration for the incorporation of cross-modal information.
Overall, we believe that this work marks a meaningful advancement in the field of DG, offering a more nuanced approach for learning invariant features.

\bibliographystyle{IEEEtran}
\bibliography{tmm}

\end{document}